\documentclass{article}

\usepackage{arxiv}

\usepackage[utf8]{inputenc} 
\usepackage[T1]{fontenc}    
\usepackage{hyperref}       
\usepackage{url}            
\usepackage{bm}            
\usepackage{mathtools}
\usepackage{booktabs}       
\usepackage{amsfonts}       
\usepackage{microtype}      
\usepackage{lipsum}
\usepackage{graphicx}
\graphicspath{ {./images/} }

\title{Towards Understanding the Interplay of Generative Artificial Intelligence and the Internet}

\author{
 Gonzalo Mart\'inez \\
  Universidad Carlos III de Madrid\\
  28911 Madrid, Spain \\
  \texttt{gonzmart@pa.uc3m.es} \\
  \And  
 Lauren Watson\\
  School of Informatics \\
  University of Edinburgh \\
  \texttt{lauren.watson@ed.ac.uk}
  \And
 Pedro Reviriego \\
  Universidad Polit\'ecnica de Madrid\\
  28040 Madrid, Spain \\
  \texttt{pedro.reviriego@upm.es} \\
  \And
  Jos\'e Alberto Hern\'andez\\
  Universidad Carlos III de Madrid\\
  28911 Madrid, Spain \\
  \texttt{jahgutie@it.uc3m.es} \\
  \And
  Marc Juarez\\
  School of Informatics \\
  University of Edinburgh \\
  \texttt{mjuarez@inf.ed.ac.uk}
  \And
  Rik Sarkar\\
  School of Informatics \\
  University of Edinburgh \\
  \texttt{rsarkar@inf.ed.ac.uk}
}

\begin{document}
\maketitle
\begin{abstract}

The rapid adoption of generative Artificial Intelligence (AI) tools that can generate realistic images or text, such as  DALL-E, MidJourney, or ChatGPT, have put the societal impacts of these technologies at the center of public debate. These tools are possible due to the massive amount of data (text and images) that is publicly available through the Internet. At the same time, these generative AI tools become content creators that are already contributing to the data that is available to train future models. Therefore, future versions of generative AI tools will be trained with a mix of human-created and AI-generated content, causing a potential feedback loop between generative AI and public data repositories.
This interaction raises many questions: how will future versions of generative AI tools behave when trained on a mixture of real and AI generated data? Will they evolve and improve with the new data sets or on the contrary will they degrade? Will evolution introduce biases or reduce diversity in subsequent generations of generative AI tools? What are the societal implications of the possible degradation of these models? Can we mitigate the effects of this feedback loop?
In this document, we explore the effect of this interaction and report some initial results using simple diffusion models trained with various image datasets. Our results show that the quality and diversity of the generated images can degrade over time suggesting that incorporating AI-created data can have undesired effects on future versions of generative models.

\end{abstract}

\section{Introduction}
\label{sec:introduction}

Generative AI tools like DALL-E for generating images from text descriptions or ChatGPT for generating conversation-like answers to text queries, have had a record-breaking adoption, gaining millions of users in a span of a few months. In addition, the popularity of these tools has fostered the development of a myriad of other generative models such as MidJourney, Stable Diffusion or Leonardo AI for image generation~\cite{texttoimageSurvey}, and LLaMA, Alpaca or Bard for natural language generation~\cite{GenerativeAIModels}. The rapid progress and adoption of these tools indicates that this is only the beginning of an era where generative models will play an instrumental role in content creation. 

A fundamental element for the continuous improvement of AI models is the use of massive training datasets from which complex models can be trained. For example, image-generation models currently rely on datasets like LAION5B which has more than five billion captioned images~\cite{LAION5B}, while language models use even larger datasets~\cite{ROOTS_dataset,LLM-datasets}. The availability of a large amount of high-quality training data is one of the major challenges in the training of accurate machine learning models~\cite{DataScarcity}. For text and images, the data is in many cases extracted from the Internet with crawlers that automatically collect billions of images and texts~\cite{LAIN400M,LLM-datasets}. 

In the near future, when new datasets are constructed by crawling the Internet, many of the images or text downloaded will have been created by generative AI tools. This means that the new training sets will be most likely significantly different from the ones we have today. It can be argued that AI-generated content could be identified and removed from the training dataset. However, this does not seem to be straightforward~\cite{DetectingGenAIImages} and may get harder as more and more generative AI models become available. Therefore, it seems likely that future generative AI models will be trained with data generated from previous AI models. In short, there will be a feedback loop where AI-generated content is used to train the next AI models that will in turn generate data used in the training of the following AI models and so on.

This opens a number of questions on how this may impact the evolution of generative AI models. On the one hand, using artificially generated samples for training can be beneficial and is indeed used in some applications where real training data is scarce~\cite{DatAugmentation} while on the other hand, it can also lead to degradation. In our case, the situation is much more complex as there will be a mixture of real and AI-generated data and also a feedback loop that can lead to more subtle effects over time. Indeed feedback loops are well known in control theory to be prone to amplify undesired effects and cause unstable behaviour that can lead to the collapse of a system in the long run~\cite{FeedbackControlTheory}. This has already been observed on the Internet with Recommender systems~\cite{FeedbackRecommender,DegenerationRecommender} and the same could occur for generative AI.

One possibility could be to try to break the loop and avoid using AI-generated data when training newer AI models. However, this would require tools that are capable of detecting AI-generated content something that seems challenging. Even if in some cases, it may be viable to detect the content generated by a given AI model~\cite{DetectingGenAIImages}, there are a myriad of tools and models that generate content and many more are coming. Therefore, it would be playing cat and mouse continuously. Moreover, AI-generated content can also be manipulated or combined, for example, AI-generated images may subsequently be edited making the detection even more complex. Therefore, it seems reasonable to assume that AI-generated content will be present in future massive training datasets extracted from the Internet.   

The potential issues of the feedback loop can be illustrated with a few simple cases. For example, many AI image generators, at least on their first versions, were unable to reproduce some elements, like the hands of persons, and created distorted or unnatural hands. If those images were used to train a newer model, would the model tend to reproduce the errors on the hands even if the newer model was now capable of drawing realistic hands? More generally, will the biases or limitations of previous AI models become a burden for newer models? Similarly, it is interesting to see if generative AI models can capture the diversity of the dataset they were trained with or if they can only generate a subset of those patterns thus creating less diverse content. If this does happen, will this be accumulative or stop at some point in time? Conversely, as the AI models evolve, would the generated content resemble the original dataset, will fidelity be gradually lost as part of the evolution? These are just a few open questions of the potential implications of the feedback loop created by using training datasets taken from the Internet.

In this article, we try to gain some initial understanding of this interplay between generative AI models and the Internet extending our preliminary work in \cite{martínez2023combining}. In more detail, we use AI image generators and a few simple datasets to model the interactions and run different experiments to evaluate how the AI generators could evolve over time. In doing so, we assume a worst case in which the training set for a given version of the AI tool is generated entirely by the previous version of the AI tool. This allows us to amplify the effects of the interaction. Then the generated images for each version are evaluated in terms of fidelity and diversity compared to the original dataset. The simulation results show how for several datasets diversity and fidelity degrade over time while for others, there is an initial degradation followed by a stabilization.    

The rest of this article is organised as follows: Section~\ref{sec:Preliminaries} reviews diffusion models for image generation and the metrics used for evaluating AI-generated images. Section~\ref{sec:methodology} overviews the methodology and datasets used for the simulation experiments conducted, which are shown in Section~\ref{sec:evaluation}. Related work is discussed in Section~\ref{sec:relatedwork}. Finally, Section~\ref{sec:conclusions} concludes this work with a main summary of its findings and conclusions.

\section{Preliminaries}
\label{sec:Preliminaries}

Generative models differ from classical discriminators or classifiers as their goal is to estimate the probability distribution from a set of samples so as to be able to generate samples that correspond approximately to that distribution. Generative models are not only applied to images but also to other complex data formats such as audio, 3D images, or even chemical molecules \cite{VAEMolecules}. Diffusion models represent one class of generative models alongside other architectures that tried to achieve the same goals, like Variational Autoencoders (VAEs) \cite{VAEs} and Generative Adversarial Networks (GANs) \cite{GANs},  however the latter are still competing with diffusion models in quality.

\subsection{Diffusion models}

Diffusion models \cite{DiffusionModelsOrigin}\cite{DiffusionModelsImprovement} are the state-of-the-art architecture of generative models which have achieved unprecedented milestones in generation quality and diversity that previous models could not achieve. They represent a general architecture with a variety of different variations available, however all are based on two phases: the forward phase and the backward phase.

During the first phase, the model gradually introduces Gaussian noise into the images until they degrade to random noise. In the second phase, the model tries to reverse the process. In other words, the model learns how to remove the noise of the image, going from an image of random noise (it can be seen as a random seed) to the real image by gradual steps, removing some noise in each step. To learn this task, the model is trained using the data that has been generated in the first phase, so that it can then be applied during inference.

The first process can be intuitively understood as the preparation of the dataset so that the neural network can learn how to turn the process around \cite{weng2021diffusion}\cite{karagiannakos2022diffusionmodels}. This can be specified as a Markov chain where each step only depends on the previous one. Therefore we can calculate it as a chain, see equation~\ref{eq:Diffusion_Forward}, where \(X_0\) is the original image, \(t\) is the number of steps taken and \(X_t\) is the final image. In each of these steps Gaussian noise is introduced, as can be seen in equation~\ref{eq:Diffusion_Noise}, where $\beta$ is the hyperparameter controlling the scale of noise introduced during the process.

\begin{equation}
   \label{eq:Diffusion_Forward}
    q(x_{1:T}|x_{0})=\prod_{t=1}^{T}q(x_{t}|x_{t-1})
\end{equation}
\begin{equation}
   \label{eq:Diffusion_Noise}
    q(x_{t}|x_{t-1})=\mathcal{N}(x_{t};\sqrt{1-\beta_{t}}x_{t-1},\beta_{t}I)
\end{equation}

On the other hand, the reverse procedure is a highly complex process where we will use a model to reverse phase 1. This process is also a Markov chain as shown in equation \ref{eq:Diffusion_Reverse}. 
The aim is to approximate the inverse conditional probability of the previous process $p_\theta(\mathbf{x}_{t-1} \vert \mathbf{x}_t)$ using a model defined by equation \ref{eq:Diffusion_Aproximation}, where \(\boldsymbol{\mu}_\theta(\mathbf{x}_t, t)\) and \( \boldsymbol{\Sigma}_\theta(\mathbf{x}_t, t))\) are the mean and the covariance matrix respectively\cite{weng2021diffusion}\cite{karagiannakos2022diffusionmodels}.

\begin{equation}
   \label{eq:Diffusion_Reverse}
    p_\theta(\mathbf{x}_{0:T}) = p(\mathbf{x}_T) \prod^T_{t=1} p_\theta(\mathbf{x}_{t-1} \vert \mathbf{x}_t)
\end{equation}
\begin{equation}
   \label{eq:Diffusion_Aproximation}
    \\p_\theta(\mathbf{x}_{t-1} \vert \mathbf{x}_t) = \mathcal{N}(\mathbf{x}_{t-1}; \boldsymbol{\mu}_\theta(\mathbf{x}_t, t), \boldsymbol{\Sigma}_\theta(\mathbf{x}_t, t))
\end{equation}

One of the main problems with diffusion models when compared to previous models was the time needed to generate new data \cite{TrilemmaGeneration}. This is caused by the fact that a Markov process has to be simulated at each generation step, which greatly slows down the process. Several alternatives to this process have emerged to try to speed it up. One of them has been the Diffusion Implicit \cite{DiffusionImplicit} models that allow replacing it with generative processes other than Markov, which are deterministic and speed up the generation process to a great extent.

\subsection{Metrics}
\label{sec:metrics}
In this section, we describe the metrics that we have used to measure the quality of the images generated by the different versions of the generative models. In particular, the metrics try to capture the fidelity of the generated images with respect to the original images and their diversity which represents their ability to cover all the types of images that are present in the original dataset. Most of the conventional metrics to measure the quality of generative AI images rely on the features extracted by the inception model~\cite{Inception} and try to capture the similarity of the generated images to those of the original dataset. However, those features are only relevant for color images of significant size and are not applicable to small black-and-white images, such as those found in the MNIST dataset. Therefore we will use two groups of metrics, one for small images and the other for larger images.

\subsubsection{Metrics for Small images}

For small images for which the inception model is not applicable, we will measure the fidelity of the generated images by training a classifier on the original dataset and measuring its accuracy on the generated images for each version. The rationale is that if the generated images are similar to the original ones, the classifier should have good accuracy. Instead, if the generated images deviate from the original ones, the classifier trained with the original dataset will have lower accuracy. On the other hand, to support the previous measure we have used Cross-Entropy. This will allow us to analyse the confidence with which the classifier determines its predictions and therefore the fidelity from another angle. If the classifier is not very sure of the prediction it will spread the confidence values over several classes, increasing the entropy. Otherwise, if it is very sure of the decision, the prediction will have a very low value. Lastly, unfortunately we have not been able to measure diversity in an efficient way with the help of the classifier because it is a more complex measure.

\subsubsection{Metrics based on Inception}

There are a number of metrics based on the inception model:

\textbf{Fr\'echet Inception Distance (FID):} This is the most widely used metric to quantify how similar the images created by a generative model are to those in the training dataset~\cite{FID}. FID is computed as follows, first a set of synthetic images is generated. Then the inception model~\cite{Inception} is run on both the training real dataset and the synthetic dataset and the features of the model are extracted. The model is widely used due to its strong performance for feature extraction. Once the features are obtained in the feature space of the inception model, we calculate the mean and covariance of the two image distributions (real and synthetic). Finally, we evaluate the distance between the two datasets $X$ and $Y$ using the following equation:

\begin{equation}
\label{eq:distanceFID}
d(X, Y) = \Vert\mu_X-\mu_Y\Vert^2 + \mathrm{Tr}\left(\Sigma_X+\Sigma_Y - 2 \sqrt{\Sigma_X \Sigma_Y}\right)    
\end{equation}
where $X$ is the original dataset and Y is the synthetic dataset.

The lower the FID distance is, the closer the synthetic dataset $Y$ to the real dataset $X$. This metric will allow us to better analyse the degradation as the comparison is done at the level of features, for example in the flowers dataset: petals, wings, etc. This will tend to capture the main elements perceived by humans.

\textbf{Precision and Density:} These two metrics also capture the fidelity of the generated images and can complement the Fr\'echet distance Inception as they allow us to analyse other essential areas of the generation models that are not covered by it \cite{PrecisionRecall}\cite{DensityCoverage}. Fidelity can be defined as the similarity of the generated image distribution to the original. To calculate this, an area formed from the features extracted by the inception model is constructed. Then it is checked the percentage of images contained in each area: Precision is the percentage of generated images contained in the area of the original dataset. The formula for this is shown in \ref{eq:Precision}. Density is another very close metric, with the advantage that it has protection against outliers. The use of a comparison of the two will give a better picture of the fidelity.

\begin{equation}
\label{eq:Precision}
\text{Precision} \coloneqq \frac{1}{M} \sum^{M}_{j=1} 1_{Y_{j}\,\in\, \text{manifold}(X_{1},...,X_{N})}
\end{equation}

\textbf{Recall and Coverage:} These two metrics capture the diversity of the generated metrics \cite{PrecisionRecall}\cite{DensityCoverage}. 
Diversity can be defined as the capability of the generative model to generate all the different classes or variations that existed in the original dataset. Recall is calculated by the percentage of original images per area of the generated dataset. The formula for this is shown in \ref{eq:Recall}. Coverage it is a metric very similar to recall with the advantage of being protected to outliers.

\begin{equation}
\label{eq:Recall}
\text{Recall} \coloneqq \frac{1}{N} \sum^{N}_{i=1} 1_{X_{i}\,\in\, \text{manifold}(Y_{1},...,Y_{N})}
\end{equation}

\section{Methodology and Model for Interaction}
\label{sec:methodology}

In this section, we describe the interaction model and the methodology used in the experiments.

\subsection{Interaction Model}
\label{sec:interaction}

\begin{figure}[h]   
    \centering
    \includegraphics[scale=0.5]{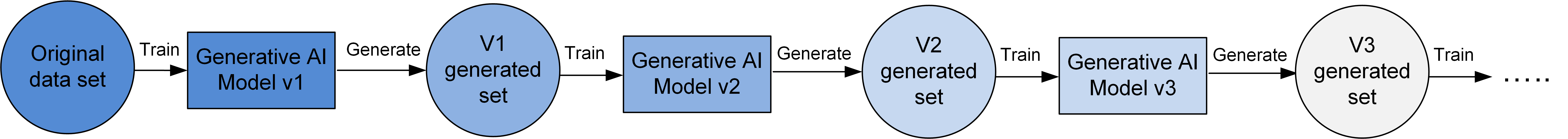}
    \caption{\textbf{Interaction model for the evolution of generative AI}}
    \label{Fig:InteractionModel}
\end{figure}

In our simulations, each version of the generative model is trained with a dataset composed of elements generated with the previous version of the generative AI model. This is illustrated in Figure~\ref{Fig:InteractionModel} for the first generations. In each generation, the size of the training set is the same as the size of the original dataset used for training. This ensures that the training time is similar in each generation. 

This interaction model corresponds to the strongest feedback possible where all the content scraped from the Internet has been generated with the latest version of the generative AI tool. Clearly, this is not realistic as the content on the Internet would be more diverse and include the non-AI generated content and content generated from several generative AI tools and for each of them for several versions. However, considering this strong interaction model allows us to analyze a worst-case scenario that should amplify the effects of the feedback loop and at the same time keep the simulations manageable. The study of more complex interaction models with datasets that grow over time with content generated from different tools and versions is left for future work.

\subsection{Methodology}

\subsubsection{Diffusion Model}

For the experiments, we have considered two types of diffusion models: First a diffusion implicit diffusion model for the flowers and Birds dataset experiments to help us with their fast sampling. On the other hand, for the MNIST experiments we have considered using a standard diffusion model with classifier-free diffusion guidance \cite{ho2022classifier} which is one of the most advanced guidance methods.

\subsubsection{Datasets}
\label{sec:datasets}

The following datasets have been used in the simulation experiments:

\begin{itemize}
    \item \textbf{MNIST~\cite{deng2012mnist}:} This is the well-known digit dataset. It contains 60,000 thousand 28x28 pixels digit images in black and white. 
    \item \textbf{Oxford 102 Flower \cite{Nilsback08}:} This dataset is made up of 8,189 color images of 102 different types of common flowers in the United Kingdom. The flower images have different sizes with most having approximately 700x500 pixels. This is a much more complex dataset than MNIST both in terms of image size and number of classes. 
    \item \textbf{Caltech-UCSD Birds-200-2011 \cite{WahCUB_200_2011}:} This is composed of 11,788 color images of 200 different types of birds. The bird images have different sizes with most of them having approximately 500x300 pixels. The image complexity is similar to the flowers dataset but with approximately twice the classes.  
\end{itemize}

\section{Simulation experiments and Evaluation}
\label{sec:evaluation}

The experiments first use the interaction model described in section~\ref{sec:interaction} to create nine generations of the diffusion model and the corresponding datasets. Then the relevant metrics among those discussed in section \ref{sec:metrics} are computed for each generation. The code along with the generated datasets is available at a public repository\footnote{https://github.com/gonz-mart/Towards-Understanding-the-Interplay-of-Generative-Artificial-Intelligence-and-the-Internet}. The experiments have been run in Google Colab platform using an A100 GPU. The time required to train the diffusion models is the largest part of the computing time and was approximately 90 minutes for the MNIST and 4 hours for the flowers and birds datasets respectively.

This section summarizes the results of the experiments analyzing both the quantitative metrics discussed in section \ref{sec:metrics} and also showing some examples of the images generated to visualize the effects.

\subsection{MNIST: Guided Diffusion on simple datasets}

In this first part, we present the results when using a denoising diffusion probabilistic model with a Classifier-Free Diffusion Guidance on the simplest dataset of the ones we consider: MNIST. The simplicity of the images on this dataset combined with a reduced number of classes and the use of guidance should improve the quality of images generated. The influence of the guidance is evaluated using different settings for this parameter. As discussed before, most of the commonly used metrics for generative models can not be used for MNIST as they are based on the Inception model that requires larger color images. Instead, a classifier is trained with the original dataset and then used to evaluate the generated images at each generation. The idea is that if the digits start to deviate significantly from the original ones, the accuracy of the classifier will drop. This would give an indication of the fidelity of the images. Additionally cross-entropy is also used to assess the confidence of the classifier. 

\begin{figure}[ht]
    \centering
    \begin{minipage}[t]{0.55\textwidth}
        \centering
        \includegraphics[width=\textwidth]{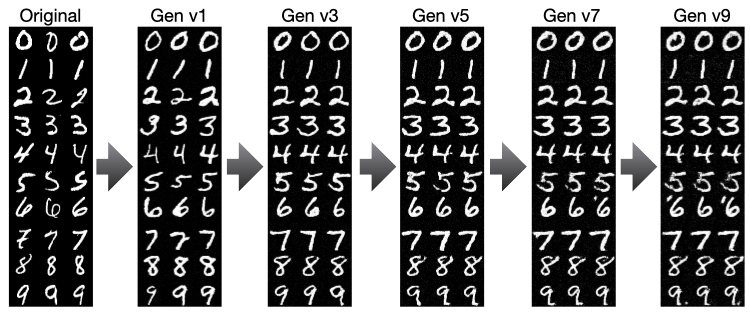}
        \textbf{Guidance: 1}
    \end{minipage}\hfill
    
    \begin{minipage}[t]{0.55\textwidth}
        \centering
        \includegraphics[width=\textwidth]{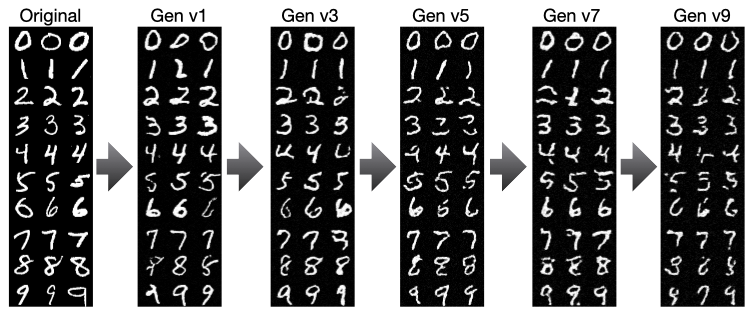}
        \textbf{Guidance: 0.1}
    \end{minipage}

    \begin{minipage}[t]{0.55\textwidth}
        \centering
        \includegraphics[width=\textwidth]{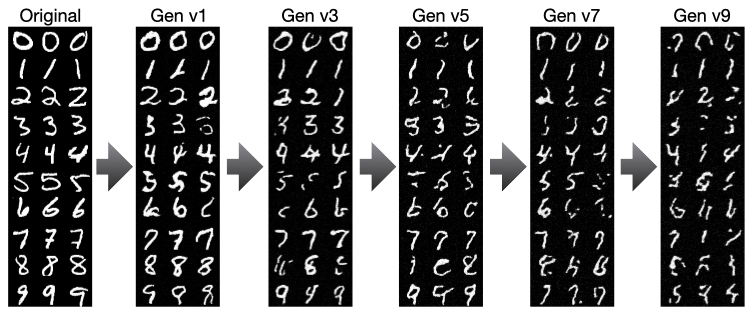}
        \textbf{Guidance: 1e-10}
    \end{minipage}

    \centering
    \begin{minipage}[t]{0.33\textwidth}
        \centering
        \includegraphics[width=\textwidth]{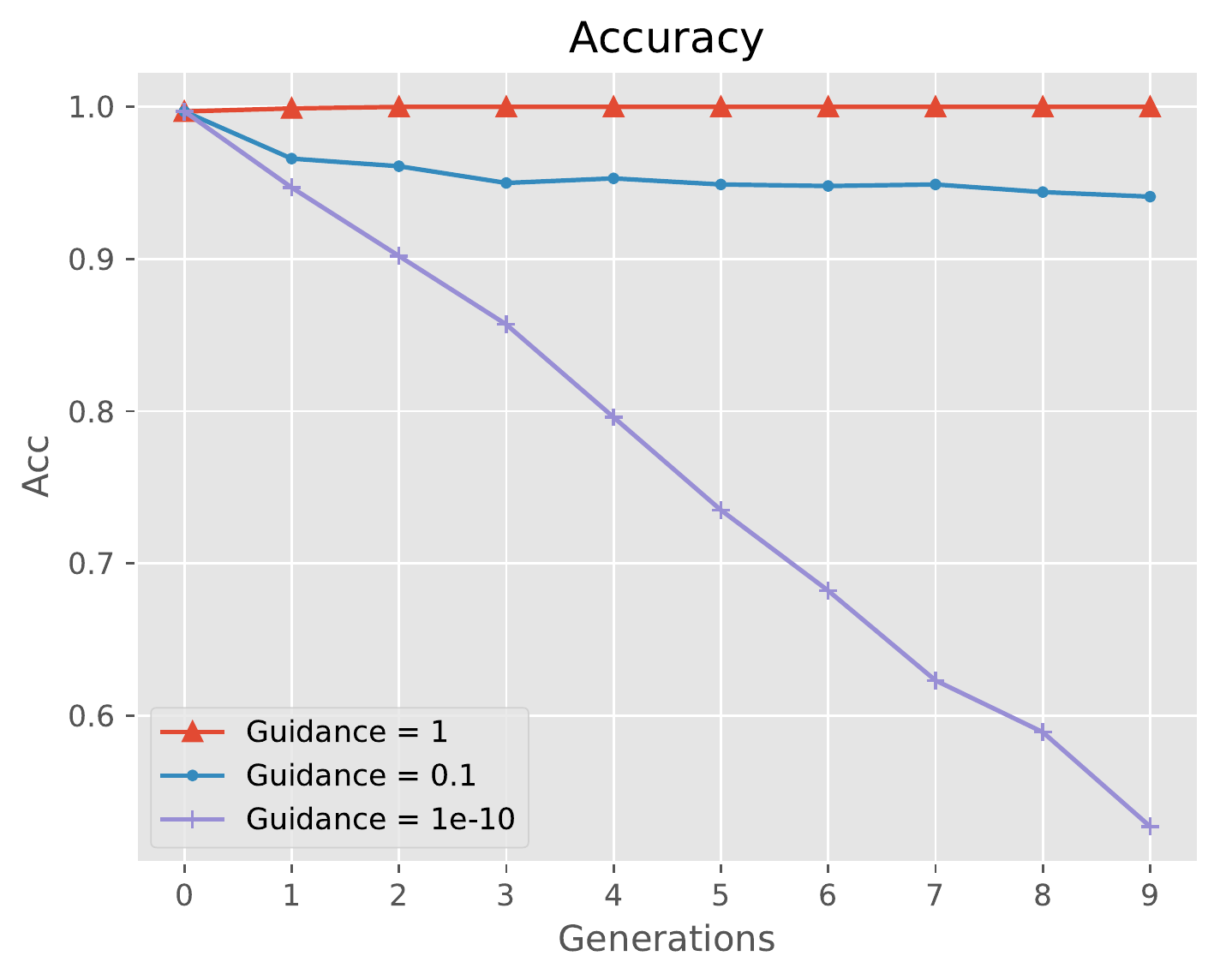}
        \textbf{a)}
    \end{minipage}\hfill
    \begin{minipage}[t]{0.33\textwidth}
        \centering
        \includegraphics[width=\textwidth]{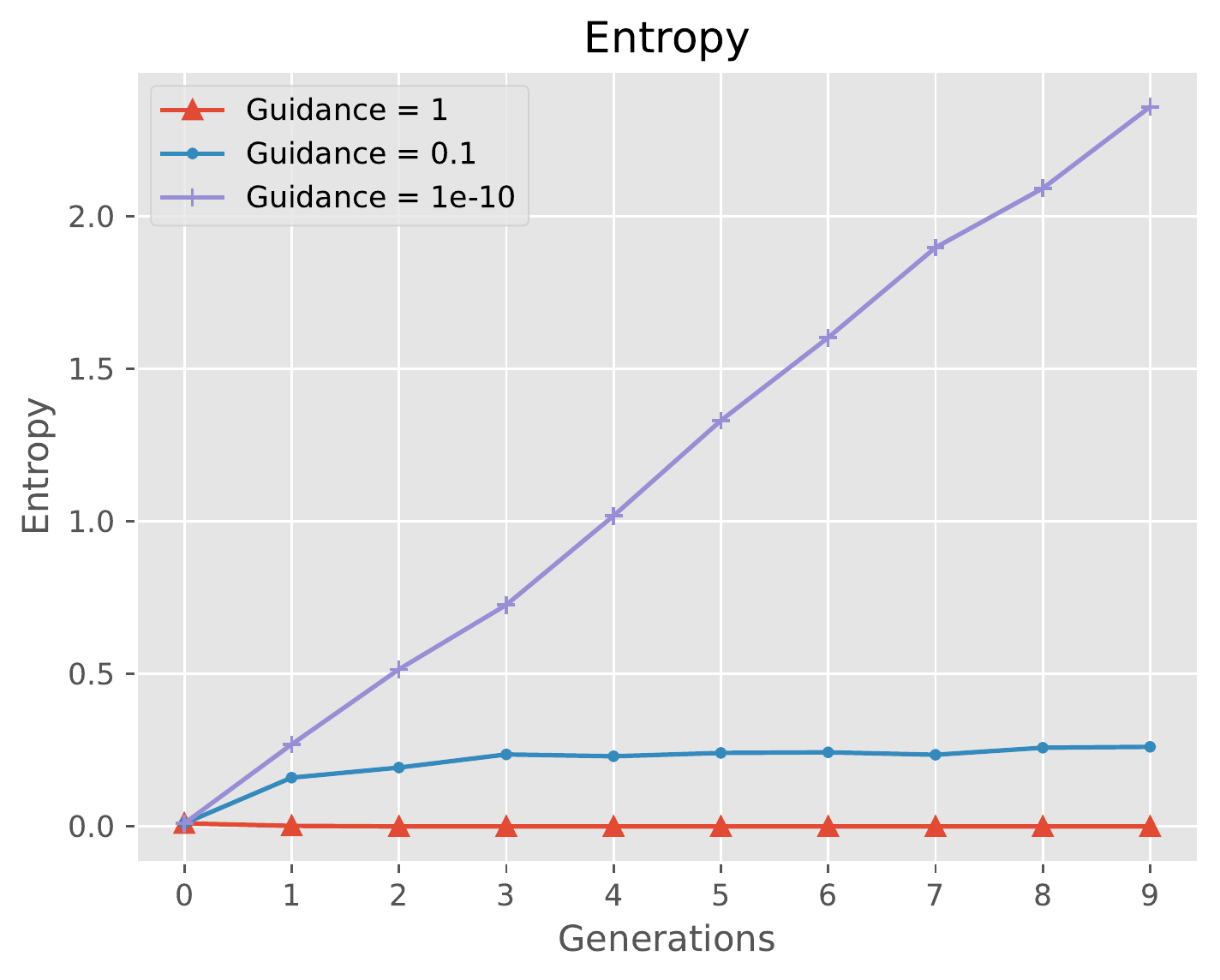}
        \textbf{b)}
    \end{minipage}
    \begin{minipage}[t]{0.33\textwidth}
        \centering
        \includegraphics[width=\textwidth]{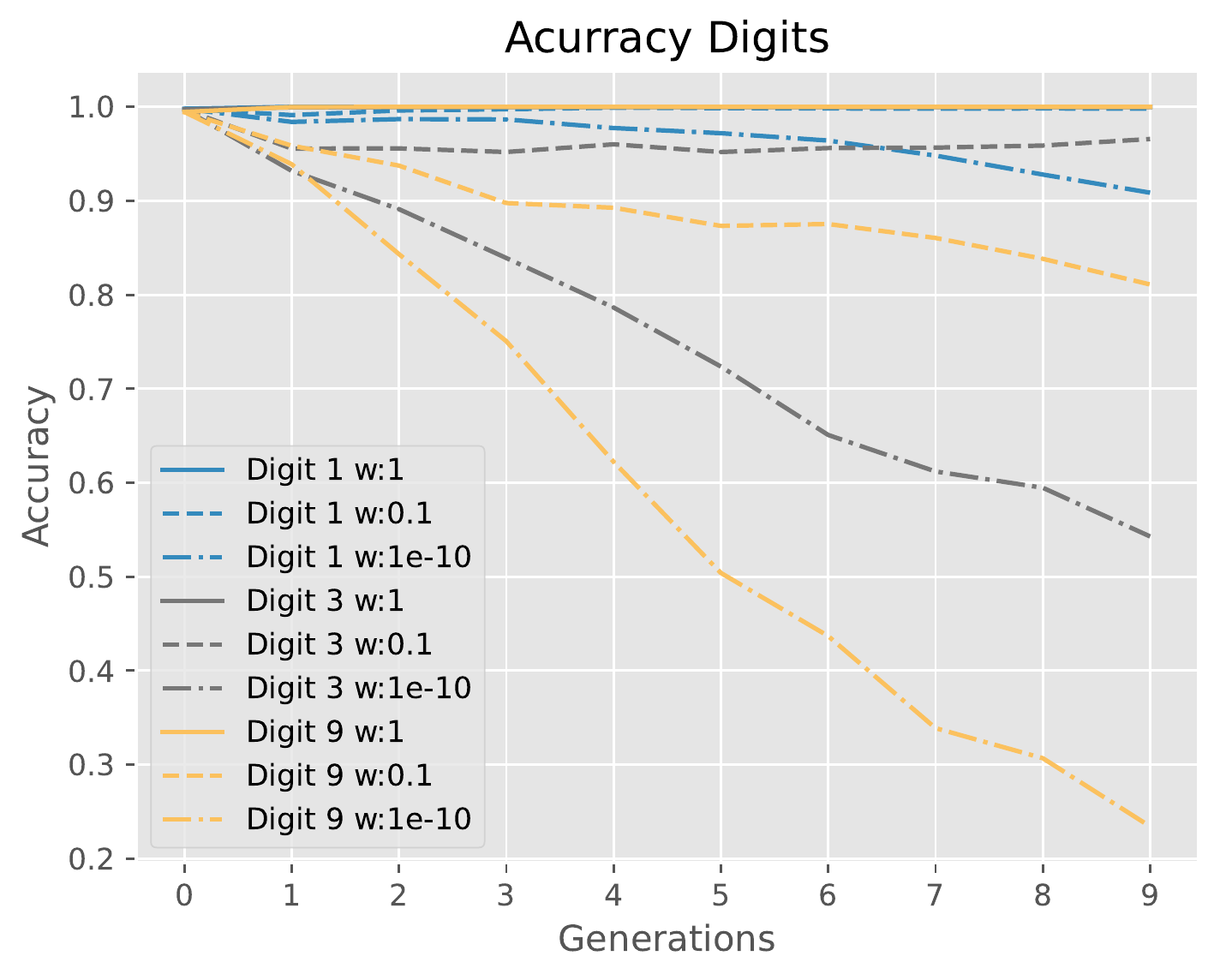}
        \textbf{c)}
    \end{minipage}
    \caption{The top plots show examples of the images generated for the MNIST dataset with different guidance values 1, 0.1, 1e-10 respectively. Plots at the bottom show relevant metrics obtained in the experiments carried out on the previously generated datasets with a classifier. Graph a) shows the results for the accuracy metric. Graph b) shows the results for the entropy and Graph c) provides the accuracy for three digits.}
    \label{fig:MNIST_Results}
\end{figure}

Figure \ref{fig:MNIST_Results} summarizes the results, in the top part sample images at different generations are shown while the metrics are placed at the bottom. Three guidance values: 1.0, 0.1, and 1e-10 have been simulated that correspond to strong, weak, and no guidance respectively. It can be observed that guidance, which indicates how closely the generated images are to the desired class, plays a key role in the evolution of the images. When a high guidance value such as 1.0 is used, it can be observed that the digits are well-defined across all the generations, in fact, they tend to be quite similar suggesting a reduction in diversity. Instead, when weaker guidance is used, for example, 0.1, it appears that there is some degeneration in the images but there is no drop in diversity. Finally, for a very low value of guidance, there is a clear degeneration in the output, and the images in some cases do not have a shape that corresponds to any digit.

The accuracy and cross-entropy are also shown at the bottom of Figure \ref{fig:MNIST_Results}. It can be clearly seen that they are consistent with the analysis of the sample images. The accuracy which we use to measure fidelity is stable and close to 100\% when guidance is 1.0. Similarly, cross-entropy is very small which shows that not only the classifier is correct but also that it is confident on its predictions. Instead, for weak guidance, there is a loss in accuracy and a larger cross-entropy over generations suggesting a degradation in the shape of the generated digits. These effects are much more apparent when no guidance is used, in this case, accuracy drops much faster and cross-entropy grows quickly. 

Finally, it is interesting to note the importance of the classes in the generation. Part c) of the Figure shows the accuracy of three digits during all the generations, which allows us to see if there has been any difference between the different classes of the data set. As can be seen in the figure, there seems to be clear differences in the degeneration, depending on the level of difficulty of the digit. Simple digits, such as one, have not been affected as much as more complex digits such as three or nine.

\subsection{Birds and Flowers: No guidance on larger datasets}

In this second part, we present the results when using a diffusion model that has no guidance in the generation when evaluated on relatively complex color images, namely the flowers and birds datasets.  

\begin{figure}[ht]
    \centering
    \begin{minipage}[t]{0.95\textwidth}
        \centering
        \includegraphics[width=\textwidth]{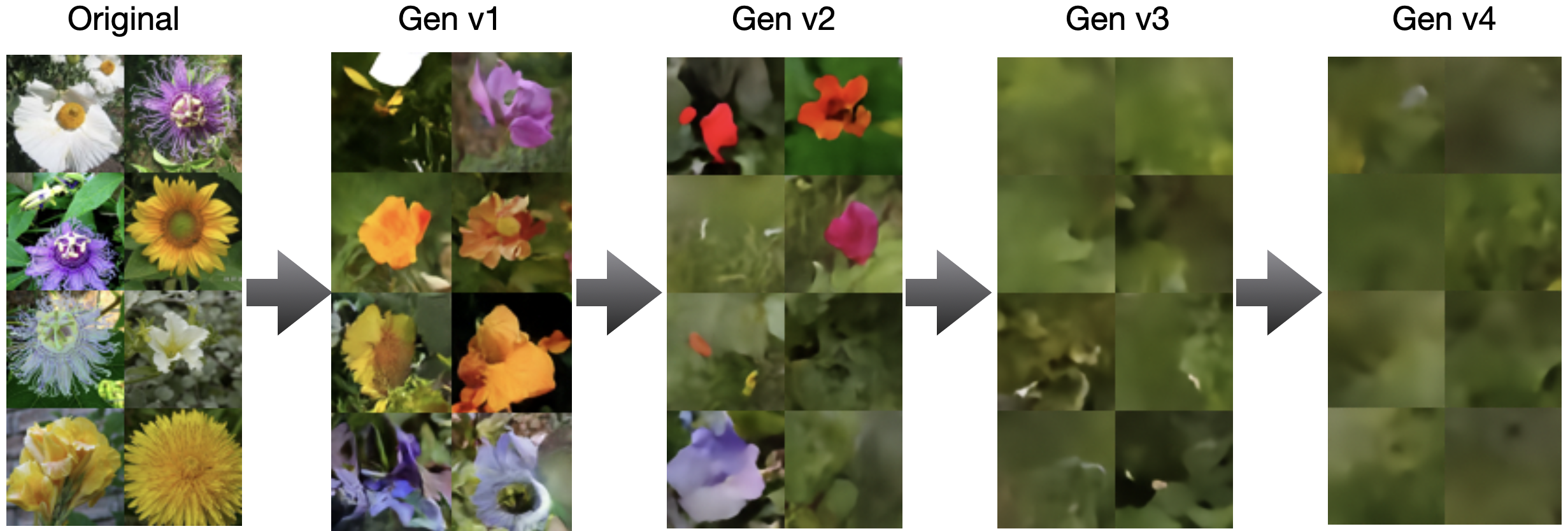}
        \textbf{a)}

    \end{minipage}\hfill
    \begin{minipage}[t]{0.95\textwidth}
        \centering
        \includegraphics[width=\textwidth]{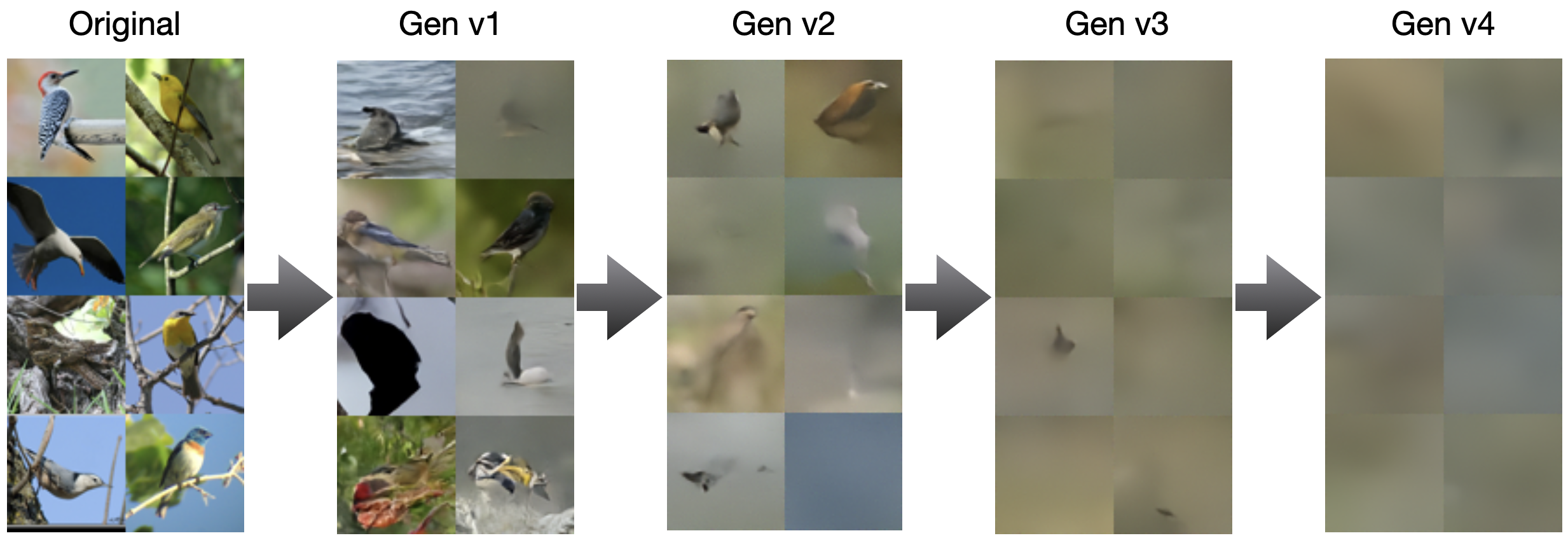}
        \textbf{b)}
    \end{minipage}
    
    \centering
    \begin{minipage}[t]{0.33\textwidth}
        \centering
        \includegraphics[width=\textwidth]{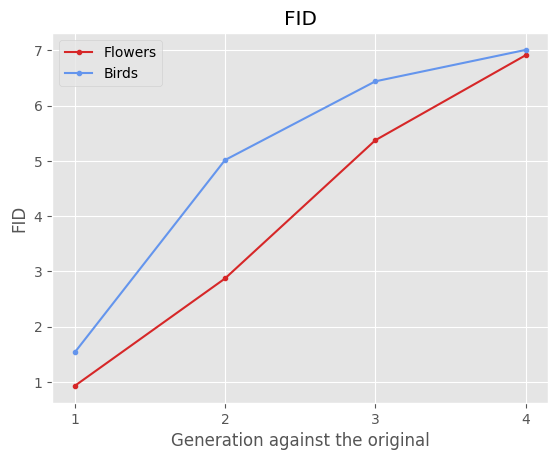}
        \textbf{c)}
    \end{minipage}
    \begin{minipage}[t]{0.33\textwidth}
        \centering
        \includegraphics[width=\textwidth]{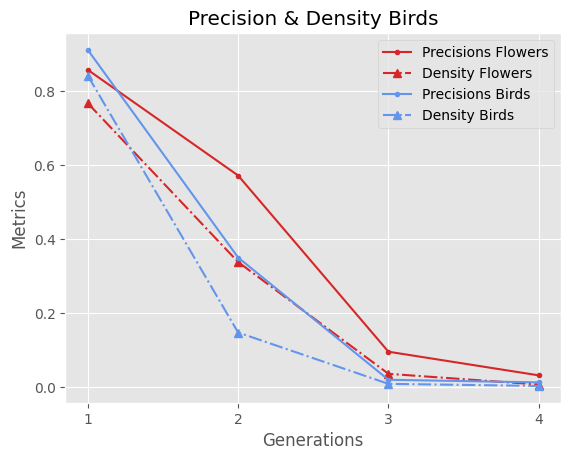}
        \textbf{d)}
    \end{minipage}
    \begin{minipage}[t]{0.33\textwidth}
        \centering
        \includegraphics[width=\textwidth]{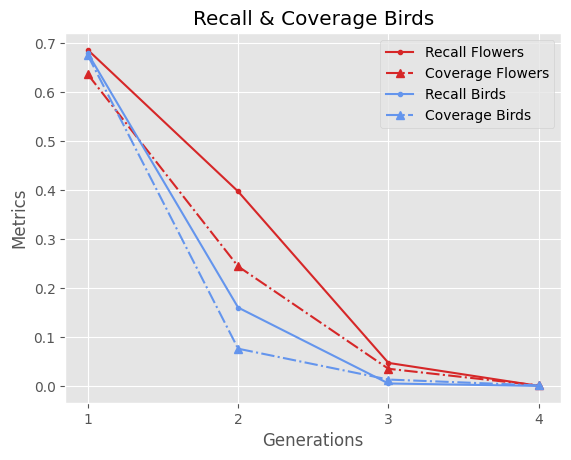}
        \textbf{e)}
    \end{minipage}
    \caption{Subfigures a) and b) show examples of the images generated with each version of the diffusion model of the  Oxford 102 Flower\cite{Nilsback08} and Caltech-UCSD Birds-200-2011\cite{WahCUB_200_2011} datasets respectively. Subfigures c), d), e) show the results obtained in the experiments carried out on the previous mencioned datasets. Graph a) shows the results of the FID metric. Graphs b) and c) measure respectively Fidelity (precision and density) and Diversity (recall and coverage). }
    \label{fig:Birds_Flowers_Results}
\end{figure}

A few samples of the images generated with each version of the diffusion model are shown in Figure \ref{fig:Birds_Flowers_Results} for flowers (a) and birds (b). As can be clearly seen, there is a degradation in each iteration of the models, first losing details in the generation, and then ending up in complete noise. After a few versions, the model is not capable of generating images that can be even recognized as flowers or birds. Therefore, in this case, evolution leads to degradation and even collapse. This can also be clearly observed when looking at the quantitative metrics that are also shown in the Figure. The Fr\'echet Inception Distance (FID) (Figure \ref{fig:Birds_Flowers_Results} (c)) increases with each version, this is consistent with what is visually observed in the figure and shows that for this diffusion model and datasets our interaction model leads to a collapse. However, we have to bear in mind that our interaction model is very strong, and on the Internet, there will be a mixture of images, not just those generated by the previous version of the generative AI tool that we are training. It is important to note that there is no progress after converging to degeneration. After reaching this point, the following models do not generate any further changes and continue to generate this noise.

Looking more in detail into the different metrics, it can be observed how the FID increases, almost linearly with each new version for the flowers while for birds, the degradation is faster in the first generations. This may be due to the birds dataset being more complex than the flowers. This may give us a clue that the degradation may be closely related to the complexity of the dataset and the inability of the model to use it. With the other metrics, the behavior is similar: we can observe how the fidelity and diversity metrics worsen with each iteration until they reach zero. This is seen in (d) for precision and density which measure fidelity and in (e) for recall and coverage which measure diversity.

\section{Related work}
\label{sec:relatedwork}

In this section we present similar studies on the contamination of data sets and similar problems that may arise from the uncontrolled generation of data on the Internet.

The first work to be highlighted is the work of Hataya et al. \cite{Datasetcorruption}, which inspired this paper and measured for the first time the possible consequences if future data sets were to be contaminated with synthetic images. His conclusions are unambiguous about the fact that only the inclusion of these synthetic images could lead to a deterioration of the quality of the data set, incapacitating the model in different tasks. As we have shown in our work. This deterioration is not only maintained but will increase in the future, so if we are not careful with the curation of the data set, this problem will be even bigger in the future.

Furthermore, it is interesting to mention that this problem in datasets is not limited to these image generation models, but can also occur in other formats such as text. This has been widely studied \cite{ContaminationTextGen} knowing that all text that has been generated is detrimental to future models and should be avoided at all costs. 

Another interesting line that we have not been able to experiment on is the bias in generation and what impact this constant generation may have. Pathways in generation is a problem that has been investigated a lot because it can lead to race, age, gender and other discriminations, and increase them over time.

However, this type of generated data is not always a problem, but can also be beneficial. An example of this is how recently some research is proposing the use of diffusion models to generate synthetic data as augmentation techniques to improve model generalisation \cite{SyntheticImproveGeneralization}. Another similar study also points in the same direction as it is the case that they use a fully synthetic data being and it proves its performance.

On the other hand, it is necessary to talk about all the work on the detection of these generated images, as it will be one of the necessary methods to maintain the quality of the datasets. If we are not able to detect these images, there will be no way to know for sure that our data samples are real. As it stands, we are in a race between detection methods and improvements in diffusion models \cite{SurveyDiffusionModels}. Because certain detection techniques are based on failures in generation, they are unlikely to be of any use in the future. Also, measures such as watermarking that allowed model owners to mark the generated images (as intellectual protection measures) are not reliable, since it is possible to disable them by using techniques to prevent their readability \cite{AvoidWatermark}.

The previously discussed works do not consider the feedback loop that can lead to accumulative effects over time. To the best of our knowledge, we were the first to investigate this issue in \cite{martínez2023combining} showing degeneration for diffusion models over generations. This paper extends our previous analysis by covering additional diffusion models and datasets as well as incorporating quantitative metrics. A recent work has followed our initial steps and considered the problem of interaction between generative models and the Internet presenting some theoretical intuition for some simple models on the causes of degeneration as well as its evaluation for natural language processing generative models \cite{shumailov2023curse}. This work complements ours and shows that the feedback can create issues in a wide range of generative models.

\section{Conclusions}
\label{sec:conclusions}

In this work we have studied the interaction of generative AI models with the Internet. As AI generated data populates the Internet it will be part of the training sets for future versions of generative AI creating a feedback loop that can have undesired effects over time. In this paper, these potential effects have been evaluated using a simple interaction model and several generative diffusion models and datasets. The results show that the interaction can lead to degeneration and also to loss of diversity. Although the results are based on a simple interaction model that should be a worst case for feedback, it confirms that the interaction should be carefully studied to understand its implications. 

This work is just a step towards the understanding of the implications of the interaction between generative AI and the Internet. Additional research is needed with more complex and realistic interaction models that use training set with a mixture of data generated with different AI models and real data. Similarly, the evaluation using more complex generative AI models and well as additional datasets is required to get a better understanding of the longterm effects of the interaction.

\bibliographystyle{unsrt}  

\bibliography{DeGeAI}

\end{document}